\documentclass[11pt]{article}

\usepackage[a4paper,margin=1in]{geometry}
\usepackage{ifpdf}

\usepackage{graphicx}
\usepackage{amsfonts}
\usepackage{amsmath}
\usepackage{amssymb}
\usepackage{amsthm}
\usepackage{algorithm}
\usepackage{algorithmic}
\usepackage{epstopdf}
\usepackage{makecell}
\usepackage{enumitem}

\usepackage[colorlinks=true,linkcolor=blue,citecolor=blue,urlcolor=blue]{hyperref}
\usepackage{cleveref}

\newcommand{\email}[1]{\href{mailto:#1}{#1}}
\newenvironment{keywords}{\par\noindent\textbf{Keywords:} }{\par}

\newtheorem{theorem}{Theorem}
\newtheorem{definition}{Definition}

\ifpdf
  \DeclareGraphicsExtensions{.eps,.pdf,.png,.jpg}
\else
  \DeclareGraphicsExtensions{.eps}
\fi

\setlist[enumerate]{leftmargin=.5in}
\setlist[itemize]{leftmargin=.5in}

\graphicspath{ {./src/} }

\begin{document}
\newcommand{\R}{\mathbb{R}}
\newcommand{\N}{\mathbb{N}}
\newcommand{\Z}{\mathbb{Z}}
\newcommand{\F}{\mathbb{F}}
\newcommand{\C}{\mathbb{C}}
\newcommand{\D}{\mathbb{D}}
\newcommand{\abs}[1]{\left\vert #1 \right\vert}
\newcommand{\norm}[1]{\left\Vert #1 \right\Vert}
\renewcommand{\Re}{\text{Re}}
\renewcommand{\Im}{\text{Im}}
\newcommand{\Part}[2]{\frac{\partial #1}{\partial #2}}
\newcommand{\argmin}{\mathop{\arg\min}}
\newcommand{\argmax}{\mathop{\arg\max}}
\newcommand{\etal}{\textit{et al. }}

\title{Harmonic Beltrami Signature Network: a Shape Prior Module in Deep Learning Framework}

\author{
    Chenran Lin\thanks{Department of Mathematics, The Chinese University of Hong Kong, Hong Kong (\email{crlin@math.cuhk.edu.hk})}
    \and
    Lok Ming Lui\thanks{Department of Mathematics, The Chinese University of Hong Kong, Hong Kong (\email{lmlui@math.cuhk.edu.hk})}
}
\date{}

\maketitle

\begin{abstract}
    This paper presents the Harmonic Beltrami Signature Network (HBSN), a novel deep learning architecture for computing the Harmonic Beltrami Signature (HBS) from binary-like images. HBS is a shape representation that provides a one-to-one correspondence with 2D simply connected shapes, with invariance to translation, scaling, and rotation. By exploiting the function approximation capacity of neural networks, HBSN enables efficient extraction and utilization of shape prior information. The proposed network architecture incorporates a pre-Spatial Transformer Network (pre-STN) for shape normalization, a UNet-based backbone for HBS prediction, and a post-STN for angle regularization. Experiments show that HBSN accurately computes HBS representations, even for complex shapes. Furthermore, we demonstrate how HBSN can be directly incorporated into existing deep learning segmentation models, improving their performance through the use of shape priors. The results confirm the utility of HBSN as a general-purpose module for embedding geometric shape information into computer vision pipelines.
\end{abstract}

\begin{keywords}
    Harmonic Beltrami Signature,
    Shape Representation,
    Image Segmentation,
    Shape Prior,
    Spatial Transformer Network
\end{keywords}

\section{Introduction}
\label{section: intro}
Image segmentation is a fundamental task in computer vision, with applications in medical image analysis, autonomous driving, and object recognition. Traditional segmentation algorithms, such as active contour models, level set methods, region growing, and graph cuts, have been widely adopted. However, they often struggle in difficult imaging conditions characterized by blurring, occlusion, low resolution, noise, and complex object boundaries. These difficulties stem largely from the absence of explicit shape prior information. Incorporating shape priors can improve segmentation accuracy and robustness by constraining the solution space to geometrically plausible outputs.

Recent developments in computational geometry have introduced shape representation techniques that address these limitations. Among these, the Harmonic Beltrami Signature (HBS) \cite{linHarmonicBeltramiSignature2022} is a well-grounded geometric descriptor. HBS encodes a 2D simply connected shape as a complex function defined on the unit disk, establishing a one-to-one correspondence between shapes and their signatures with invariance to translation, scaling, and rotation. HBS has shown strong performance in shape representation, and has been applied to image classification and segmentation tasks.

Deep learning has also produced a range of strong methods for image segmentation, using large datasets to learn mappings from images to pixel-wise labels. Representative architectures include UNet \cite{ronneberger2015u}, Fully Convolutional Network (FCN) \cite{long2015fully}, SegNet \cite{badrinarayanan2016segnet}, DeepLab \cite{chen2017deeplab,chen2016semantic, chen2017rethinking}, Pyramid Scene Parsing Network (PSPNet) \cite{zhao2017pyramid}, and Mask Region-based Convolutional Neural Network (Mask RCNN) \cite{zhao2017mask}. These models achieve strong performance in segmentation tasks, with the ability to capture fine-grained details and semantic information. However, while effective at learning visual features, they do not have explicit mechanisms for extracting or incorporating geometric shape priors. Integrating such priors into deep architectures could improve segmentation accuracy and robustness.

Motivated by the success of HBS in segmentation \cite{lin2024shape}, we introduce the Harmonic Beltrami Signature Network (HBSN), a neural network module that incorporates geometric shape information directly into the segmentation pipeline. Built on the UNet architecture and equipped with Spatial Transformer Network (STN) \cite{jaderberg2015spatial} components, HBSN predicts HBS representations from binary images, allowing deep learning models to use shape priors during training.

Our primary contributions are as follows:
\begin{enumerate}
    \item Development of HBSN, a specialized network for computing Harmonic Beltrami Signatures.
    \item Demonstration of the integration of shape priors into deep learning segmentation architectures, improving accuracy and robustness in complex scenes.
    \item Experimental evaluation of HBSN across multiple computer vision tasks, with emphasis on segmentation.
\end{enumerate}

Through experiments, we validate HBSN's ability to compute HBS efficiently and demonstrate its applicability to segmentation tasks. This work provides a systematic approach to incorporating shape prior information into deep learning models for complex visual tasks.

The rest of the paper is organized as follows:
\cref{section: related work} shows some related topics about image segmentation;
\cref{section: background} introduces some theoretical background;
\cref{section: main} explains our proposed HBSN in detail;
\cref{section: exp} reports our experimental results;
the paper is concluded in \cref{section: conclusion}, and we point out several future directions.

\section{Related works}\label{section: related work}
We review related work in several areas of segmentation relevant to HBSN.

\subsection{Traditional segmentation algorithms}
Traditional segmentation methods, such as thresholding, region growing, edge detection, and clustering, form the foundation of image segmentation but often falter in complex scenes due to the absence of explicit shape priors. Representative methods include the watershed algorithm \cite{beucher1993watershed} for images with varying complexity, curvature-based region growing \cite{leymarie1993curvature} for accurate boundary delineation, active contours \cite{kass1988snakes} as a deformable model framework, mean shift clustering \cite{comaniciu1999mean} based on non-parametric density estimation, and geodesic active contours \cite{caselles1993geometric} leveraging level sets with geometric properties.

\subsection{Shape prior segmentation algorithms}
Incorporating shape priors into segmentation enhances accuracy and robustness. Notable approaches include variational frameworks with shape priors \cite{tsai2001curve}, level set-based joint segmentation and shape integration \cite{rousson2002level}, graph cuts combined with shape priors for interactive segmentation \cite{boykov2001interactive}, geodesic matting with shape constraints \cite{bai2007geodesic}, and statistical frameworks for shape-based segmentation \cite{rathi2007shape}.

\subsection{Deep learning segmentation networks}
Recently, deep learning segmentation networks have achieved strong performance by learning complex image-to-label mappings. Key architectures include UNet \cite{ronneberger2015u} with its encoder-decoder design and skip connections for biomedical segmentation, FCN \cite{long2015fully} enabling end-to-end pixel-wise segmentation, SegNet \cite{badrinarayanan2016segnet} for efficient semantic segmentation, DeepLab \cite{chen2017deeplab} employing atrous convolution and spatial pyramid pooling for multi-scale context, and Mask RCNN \cite{zhao2017mask,ren2016faster} extending object detection with instance segmentation.

\subsection{Shape prior deep learning segmentation networks}
Integrating shape priors into deep learning has led to specialized segmentation networks, including shape-prior frameworks for object segmentation \cite{zhang2018shape}, shape-agnostic kernels for learning shape-aware features \cite{sanakoyeu2019shape}, shape-guided Siamese networks for video tracking \cite{xu2020deep}, feature regularization with shape priors \cite{liao2019deep}, and shape-constrained semantic segmentation \cite{wang2019shape}.

\subsection{Quasi-conformal theory}
Quasi-conformal theory, the theoretical foundation of our work, has been widely applied in computer vision \cite{zhang2023deformationinvariant,lyu2024spherical,guo2023automatic,lyu2023two,zhu2022parallelizable,zhang2022unifying,zhang2022new,zhang2022nondeterministic,zhang2021quasi}. Recent applications include interactive segmentation \cite{zhang2024qis}, topology-preserving segmentation networks \cite{zhang2022topology}, density-equalizing surface flattening \cite{lyu2024deq}, and diffeomorphic image registration \cite{chen2024deep}.

\section{Theoretical basis}\label{section: background}

\subsection{Quasi-conformal mapping and Beltrami equation}

A complex function $f: \Omega \subset \C \rightarrow \C$ is said to be \textit{quasi-conformal} associated to $\mu$ if $f$ is orientation-preserving and satisfies the following \textit{Beltrami equation}:
\begin{equation}\label{eq: beltrami eq}
    \Part{f}{\overline{z}} = \mu(z) \Part{f}{z}
\end{equation}
where $\mu(z)$ is a complex-valued Lebesgue measurable function satisfying $\norm{\mu}_\infty < 1$. More specifically, this $\mu: \Omega \rightarrow \D$ is called the \textit{Beltrami coefficient} of $f$
\begin{equation}\label{eq: mu def}
    \mu = \frac{f_{\overline{z}}}{f_{z}}
\end{equation}

\begin{figure}
    \begin{center}
        \includegraphics[width=5cm]{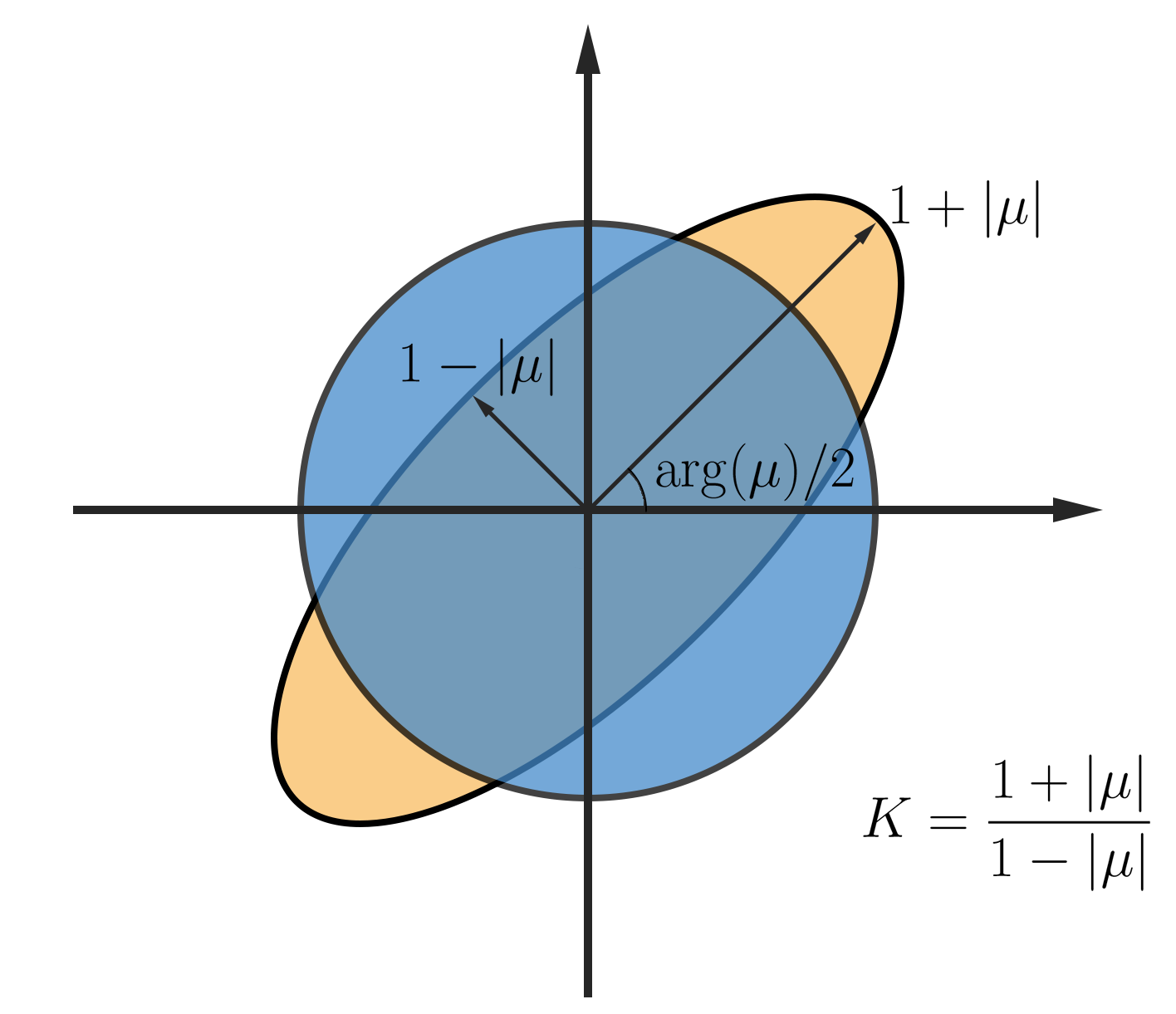}
    \end{center}
    \caption{Quasi-conformal maps infinitesimal circles to ellipses. The Beltrami coefficient measures the distortion or dilation of the ellipse under the QC map.}
    \label{fig: qc}
\end{figure}

Geometrically, a quasi-conformal map $f$ maps infinitesimal circles to ellipses with dilation $K = (1+|\mu|)/(1-|\mu|)$, and the Beltrami coefficient $\mu$ fully characterizes this conformal distortion (see \cref{fig: qc}). When $\mu \equiv 0$, $f$ is conformal.

Note that there is a one-to-one correspondence between the quasi-conformal mapping $f$ and its Beltrami coefficient $\mu$. Given $f$, there exists a Beltrami coefficient $\mu$ satisfying the Beltrami equation by equation \cref{eq: mu def}. Conversely, the measurable Riemann mapping theorem \cite{ahlfors1960riemann} states that given an admissible Beltrami coefficient $\mu$, a quasi-conformal mapping $f$ always exists associated with this $\mu$.

\subsection{HBS}
\begin{figure}
    \begin{center}
        \includegraphics[width=12cm]{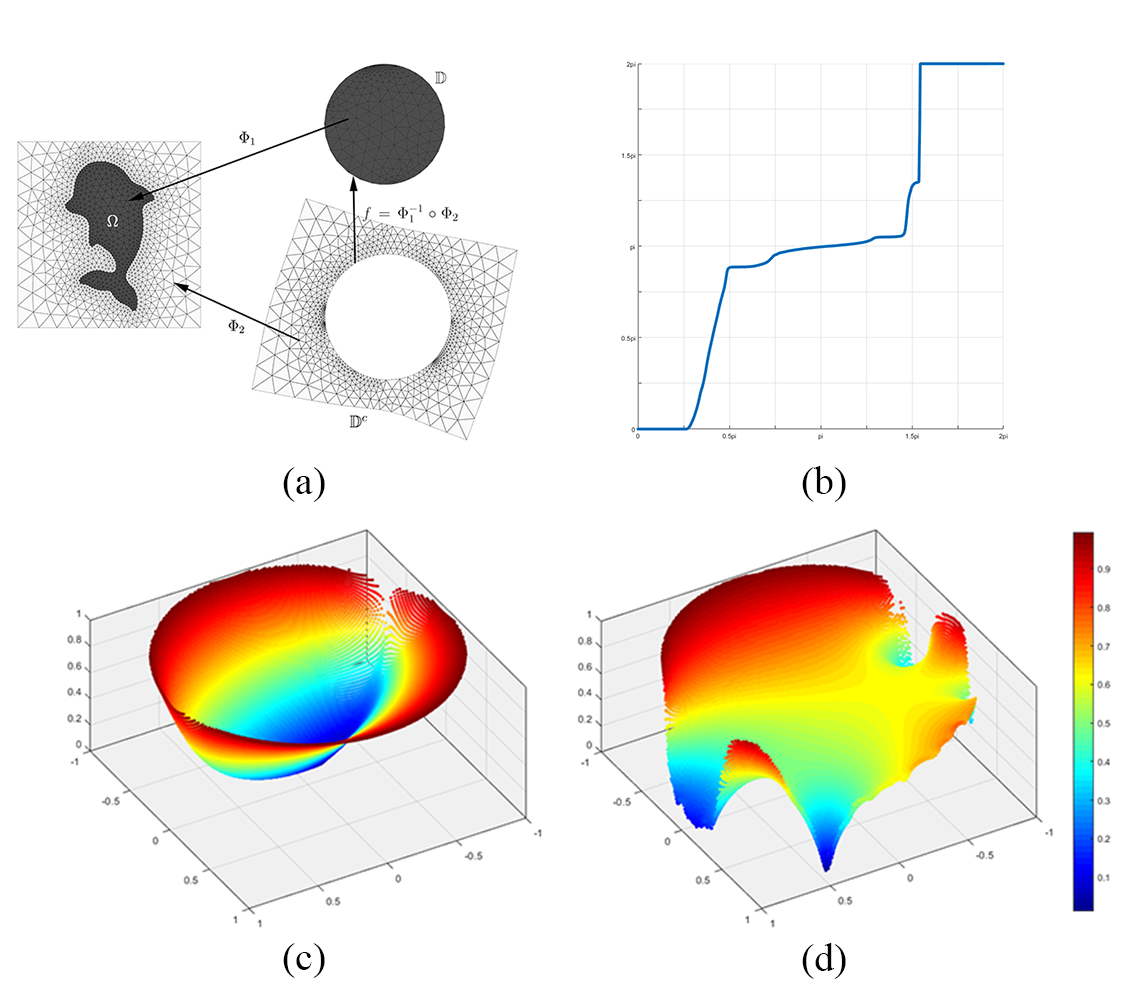}
    \end{center}
    \caption{The illustration of HBS. (a) shows the shape $\Omega$ and conformal maps $\Phi_1$ and $\Phi_2$; (b) is the conformal welding $f = \Phi_1^{-1} \circ \Phi_2$; (c) is the Harmonic extension $H$ of conformal welding $f$; (d) is the GHBS $B$ corresponding to $H$.}
    \label{fig: ghbs}
\end{figure}

Suppose $\Omega\subset \mathbb{C}$ is a Jordan domain and a quasicircle. Let $f = \Phi_1^{-1} \circ \Phi_2$ be the conformal welding of $\Omega$, where  $\Phi_1: \D \rightarrow \Omega$ and $\Phi_2: \D^c \rightarrow \Omega^c$ are the conformal mappings. The harmonic extension $H: \overline{\D} \to \C$ of $f$ is achieved by the Poisson integral on the unit circle. Then the Beltrami coefficient of $H$ is called \textit{Generalized Harmonic Beltrami Signature (GHBS)}
\begin{equation*}
    B(z):= \mu_H(z) = \frac{H_{\bar{z}}(z)}{H_z(z)}.
\end{equation*}
The process of obtaining GHBS is shown in \cref{fig: ghbs}.

When the $\Phi_2$ of GHBS satisfies $\Phi_2(\infty) = \infty$, we call this GHBS fixed at infinity, and then the equivalence relation of GHBS can be defined as
\begin{definition}
    Suppose two GHBS $B$ and $\tilde{B}$ are fixed at infinity, they are said to be {\it equivalent} if $B=\mu_H$ and $\tilde{B} = \mu_{\tilde{H}}$, where $H$ and $\tilde{H}$ are respectively the harmonic extensions of a diffeomorphism $f:\mathbb{S}^1\to \mathbb{S}^1$ and $\tilde{f} = M_1^{-1} \circ f \circ M_2$ with $M_1$ is a Mobi\"us transformation and $M_2$ is a rotation. In this case, we denote $B \sim \tilde{B}$. Also, the equivalence class of $B$ is denoted by $[B]$.
\end{definition}

We consider the space of GHBS equivalence classes $\mathcal{B} = \mathcal{B}_0 \,/ \sim$ to study the quotient space of shapes $\mathcal{S} = \mathcal{S}_0 \, / \sim_{\mathcal{S}}$, where $\mathcal{B}_0 = \{B:\mathbb{D}\to \mathbb{D} \mid B \text{ is a GHBS fixed at infinity} \}$, $\mathcal{S}_0 = \{\Omega \subset \C \mid \Omega \text{ is Jordan domain}\}$, $\Omega \sim_{\mathcal{S}} \bar{\Omega}$ iff $\bar{\Omega} = F(\Omega)$ and $F$ is composed of translation, rotation and scaling. The following theorem illustrates that GHBS is an effective representation.

\begin{theorem}\label{thm: one to one equivalence class}
    There is a one-to-one correspondence between $\mathcal{B}$ and $\mathcal{S}$. In particular, given $[B]\in \mathcal{B}$, its associated shape $\Omega$ can be determined up to a translation, rotation, and scaling.
\end{theorem}

The \textit{Harmonic Beltrami Signature (HBS)} $B$ is the unique representative of a GHBS equivalence class, which satisfies the following conditions:
\begin{equation}\label{eq: hbs conditions}
    \int_{\partial \Omega} \Phi_1^{-1}(z) dz = 0,
    \quad \arg \int_\D B(z) dz = 0,
    \quad \arg \int_\D \frac{B(z)}{z} dz \in [0, \pi).
\end{equation}

Given shape $\Omega$, its HBS $B$ is uniquely determined and is invariant under rotation, translation, and scaling. The following theorem guarantees the geometric stability of HBS:
\begin{theorem}\label{thm: stability of HBS}
    Let $B_1$, $B_2$ be two HBS and $\Omega_1$ and $\Omega_2$ be the normalized shapes associated to $B_1$ and $B_2$ respectively. If
    $||B_1 - B_2||_{\infty} < \epsilon$, then the Hausdorff distance between $\Omega_1$ and $\Omega_2$ satisfies
    \begin{equation*}
        d_H(\Omega_1,\Omega_2)
        = \max \left(
        \max_{q \in \Omega_2} \min_{p \in \Omega_1} \abs{p-q},
        \max_{p \in \Omega_1} \min_{q \in \Omega_2} \abs{p-q}
        \right)
        < \frac{2M}{\pi}\epsilon
    \end{equation*}
    for some $M>0$.
\end{theorem}

Besides, the effective reconstruction algorithm from an HBS $B$ to the original shape $\Omega$ is also proposed and proved in \cite{linHarmonicBeltramiSignature2022}. This algorithm constructs a function $G: \C \rightarrow \C$ such that $G(\D) = \Omega$ and the Beltrami coefficient of $G$ is
$$
    \mu_G = \begin{cases}
        B \text{ on } \D, \\
        0 \text{ on } \D^c.
    \end{cases}
$$

\section{Proposed network}\label{section: main}
In this section, we describe our proposed shape prior deep neural network module, called the Harmonic Beltrami Signature Network (HBSN), to compute the HBS of the given binary image containing a 2D bounded simply connected shape $\Omega$.

\begin{figure}
    \centering
    \includegraphics[width=10cm]{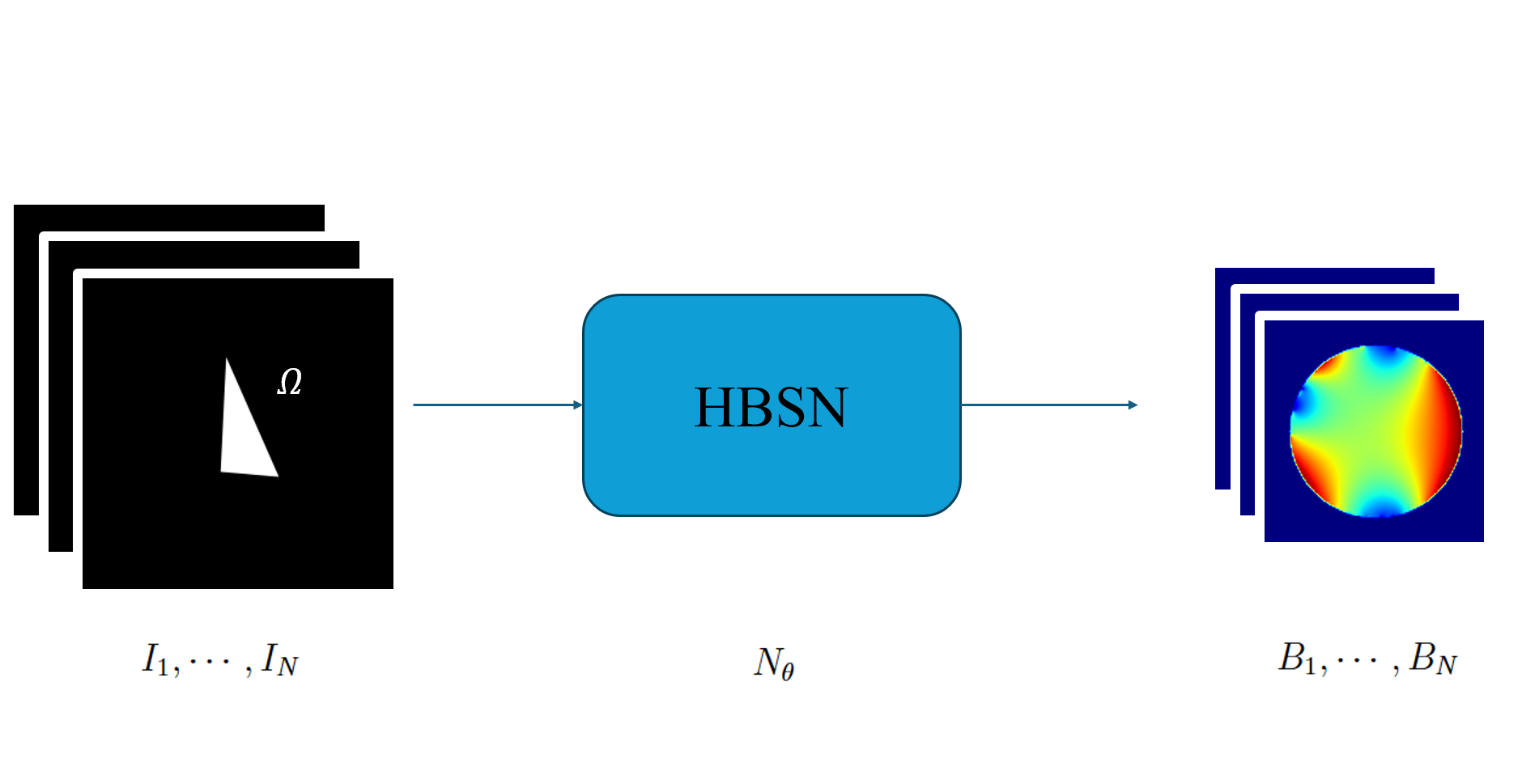}
    \caption{The illustration of HBSN $N_\theta$. It receives 2D rectangle binary images $I_1, \cdots, I_N$ with simply connected shapes and then calculates their HBS $B_1, \cdots, B_N$.}
    \label{fig: hbsn}
\end{figure}

\subsection{Problem setting}
Given a 2D binary image $I: D \to \{0, 1\}$ with a bounded simply connected shape $\Omega \subset D \subset \C$
\begin{equation}\label{eq: image with shape}
    I(x) = \begin{cases}
         & 1, \; x \in \Omega,              \\
         & 0, \; x \in D \setminus \Omega ,
    \end{cases}
\end{equation}
then the corresponding HBS $B: \D \to \C$ of shape $\Omega$ is a complex function defined on unit disk. The HBS $B$ is uniquely determined by $\Omega$ and invariant under translation, scaling, and rotation, making it a stable shape prior. Moreover, the distance between two HBS can be measured directly by the $L^2$ norm, and the HBS changes linearly with shape distortion within a reasonable range, which makes it a useful tool for quantifying shape differences. These properties of HBS have been validated in image classification and segmentation tasks \cite{linHarmonicBeltramiSignature2022,lin2024shape}.

An algorithm is also proposed in \cite{linHarmonicBeltramiSignature2022} for computing HBS, which could be summarized as follows:
\begin{enumerate}
    \item Pick clockwise boundary points $z_1, ...  z_N \in \partial \Omega$;
    \item Calculate conformal welding $f$ of $\partial \Omega$ by Zipper algorithm;
    \item Extend $f$ to Harmonic extension $H$ by Poisson Integral;
    \item Get HBS $B$ by computing the Beltrami Coefficient of $H$.
\end{enumerate}


Unfortunately, this algorithm fails to give an explicit formula from shape $\Omega$ to HBS $B$, and it is even hard to compute gradients since it has too many conditional branches and loops. This shortage also forces us to choose an indirect method in the HBS segmentation model. In this gradient-based iteration model, we achieved the final segmentation result by approximating reference HBS under some slacked constraints.

In this paper, we try to find another solution under the framework of deep learning neural networks. Suppose a binary image $I$ containing shape $\Omega$ defined as \cref{eq: image with shape}, and the HBS $B$ of $\Omega$ could be extended to a rectangle domain $D' \supset \D$ by setting the value outside unit disk to 0,
\begin{equation}\label{eq: HBS of image}
    B_I(x) = \begin{cases}
         & B(x), \; x \in \D,           \\
         & 0, \; x \in D' \setminus \D,
    \end{cases}
\end{equation}
then we call $B_I$ as the HBS of such image $I$. In the following discussion, we will not deliberately distinguish between the HBS of the image $I$ and the shape $\Omega$, so we ignore the subscript and use $B$ to refer to the HBS of image $B_I$. For the convenience of discussion, the domains $D$ and $D'$ are rectangles, and $0$ is fixed as the center of $D'$. Under discrete situations, the image and HBS could be considered as matrices $I \in M_{m \times n}(\R)$ and $B \in M_{p \times q}(\C)$. And in the experiments of this paper, we fix $m=n=256$, $p=q=128$.

Our strategy is to find a deep neural network $N_\theta: M_{m \times n}(\R) \to M_{p \times q}(\C)$ with network parameters $\theta$ such that
\begin{equation}
    \min_\theta L(N_\theta(I), B),
\end{equation}
where $L: M_{p \times q}(\C) \times M_{p \times q}(\C) \to \R$ is some loss function.

Once the network $ N_theta$ is designed and successfully trained, the HBS $B_I$ of the given image $I$ can be obtained in real-time.
Since the network is differentiable by construction, gradients can be computed via backpropagation, which makes it straightforward to couple HBSN with other neural network models.

\subsection{Overall architecture}
\begin{figure}
    \centering
    \includegraphics[width=10cm]{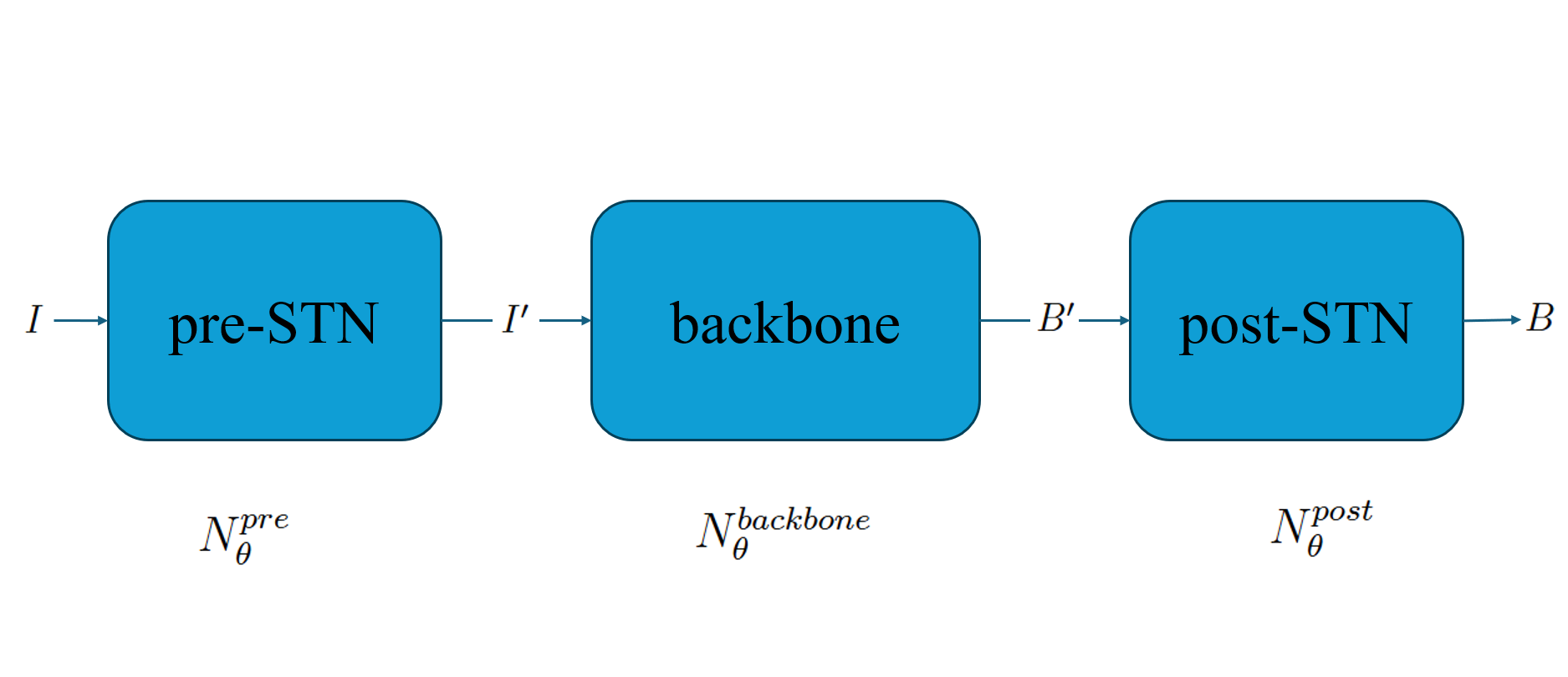}
    \caption{Overall architecture of the proposed HBSN.}
    \label{fig: overall architecture}
\end{figure}

The overall architecture of the proposed network is depicted in \cref{fig: overall architecture}.
To achieve the HBS with expected properties like uniqueness and invariance, our HBSN $N_\theta$ consists of 3 main blocks: (1) the pre Spatial Transformer Network (pre-STN) $N^{pre}_\theta: M_{m \times n}(\R) \to M_{m \times n}(\R)$, (2) the backbone $N_\theta^{backbone}: M_{m \times n}(\R) \to M_{p \times q}(\C)$, and (3) the post Spatial Transformer Network (post-STN) $N_\theta^{post}: M_{p \times q}(\C) \to M_{p \times q}(\C)$.

The pre-STN $N_\theta^{pre}$ estimates the relative space position of the given image $I$ and then returns a position normalized image $I' = N_\theta^{pre}(I)$.
The backbone $N_\theta^{backbone}$ is an encoder-decoder network, which extracts the shape features of $I'$ then reconstructs them in the form of the HBS to get $B' = N_\theta^{backbone}(I')$.
The post-STN $N_\theta^{post}$ is only responsible for adjusting the rotation of $B'$ to generate the unique output $B = N_\theta^{post}(B')$. So far, the desired HBS $B = N_\theta(I)$ is achieved through these three blocks. The detailed architecture and functions of each block will be discussed later.

We note that there is no theoretical guarantee that the output $B$ of HBSN is exactly the HBS of the given image. Nevertheless, the experimental results consistently show that $B$ is close to the ground truth and that the network generalizes well to unseen shapes.

\subsection{Backbone}
\begin{figure}
    \centering
    \includegraphics[width=\textwidth]{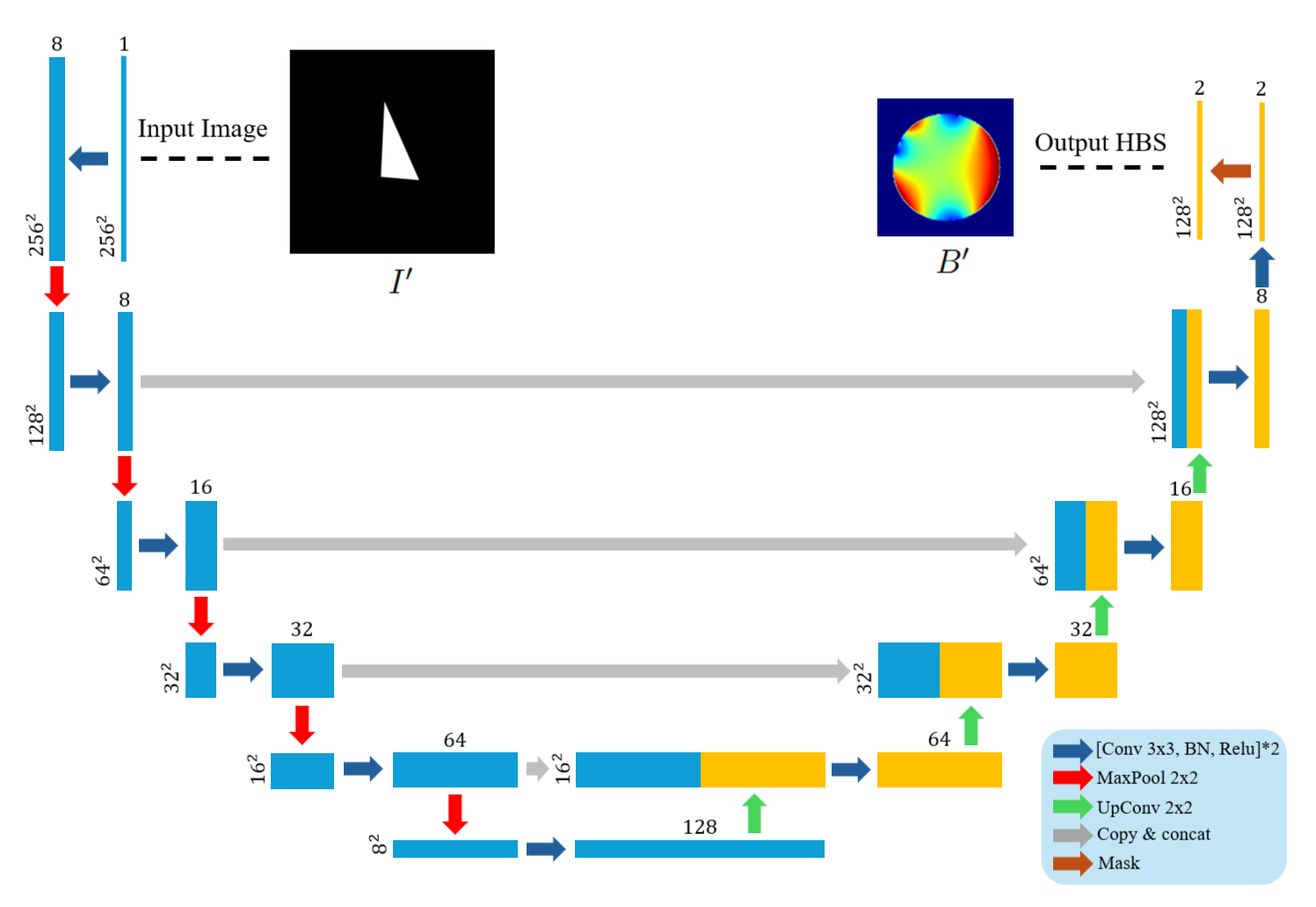}
    \caption{The architecture of the backbone.}
    \label{fig: hbsn backbone}
\end{figure}

The backbone block of our HBSN is based on UNet \cite{ronneberger2015u}, whose architecture is shown in \cref{fig: hbsn backbone}.

The encoder extracts different levels of image features. Unlike the standard UNet, we adjusted the number of channels per layer based on the properties of HBS computation from binary images. The early encoder layers typically capture texture features, but binary images contain almost no texture apart from the shape boundary. Since boundary information is the only relevant signal for HBS computation and can be extracted straightforwardly by convolutions, fewer channels are used in the early layers, as marked in the figure.
However, the global information of shape is critical to HBS. There are five times downsamples in the encoder of the backbone, which means the size of the image decreases from $256 \times 256$ to $8 \times 8$. No reliable network could be trained if using four or fewer times pooling in our experiments.

Since $I' \in M_{m \times n}(\R)$, $B' \in M_{p \times q}(\C)$, and $m,n,p,q \in \Z$ has no direct connection, an asymmetric structure is utilized in the decoder. As shown in the architecture figure, we only upsample four times and use a $128 \times 128$ image as output, reducing computation complexity.

Recall that the value of HBS is only meaningful on the unit disk and is $0$ strictly outside, so a unit disk mask module is added at the end of the backbone. We set a circular area with a radius of 50 pixels centered around the center of the output image to represent the unit disk. The output value of UNet stays unchanged if inside the mask; otherwise, it becomes $0$ and stops the gradient backpropagation there.

\subsection{Pre- and Post-STN}
\begin{figure}
    \centering
    \includegraphics[width=11cm]{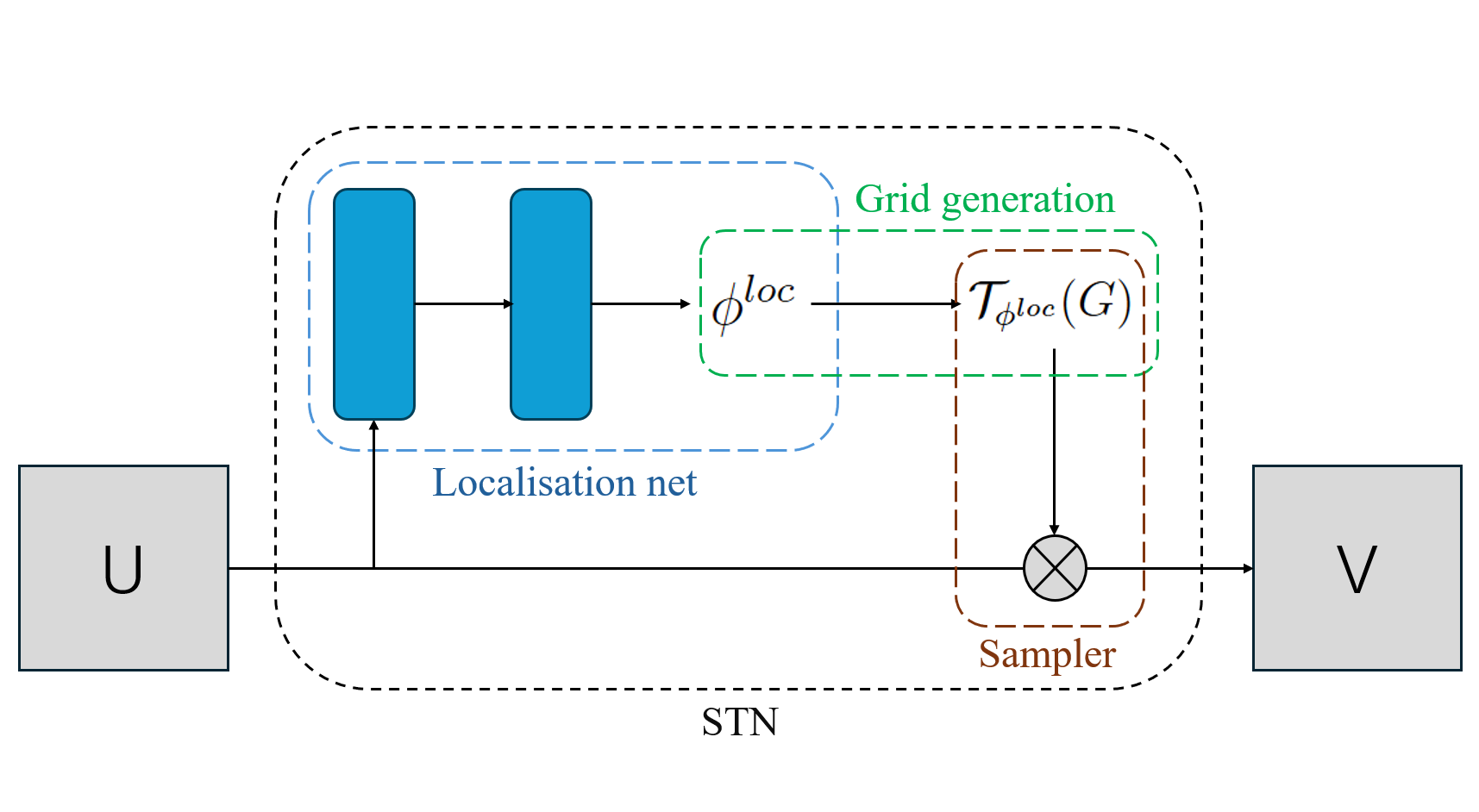}
    \caption{The architecture of the pre- and post-STN.}
    \label{fig: hbsn stn}
\end{figure}

\cref{fig: hbsn stn} illustrates the architecture of pre-STN and post-STN, both modified from STN \cite{jaderberg2015spatial}. The original STN uses a localization network to estimate a location parameter $\phi^{loc} \in \R^6$ from the given image, then applies an affine transformation to the original image according to
\begin{equation}\label{eq: base STN}
    \begin{pmatrix}
        x_i^s \\
        y_i^s
    \end{pmatrix} = \begin{pmatrix}
        \phi^{loc}_1 & \phi^{loc}_2 & \phi^{loc}_3 \\
        \phi^{loc}_4 & \phi^{loc}_5 & \phi^{loc}_6
    \end{pmatrix} \begin{pmatrix}
        x_i^t \\
        y_i^t \\
        1
    \end{pmatrix},
\end{equation}
where $(x_i^s, y_i^s)$ and $(x_i^t, y_i^t)$ are the $i$-th pixels of source image $I$ and target image $I'$. The dimensions of the location parameters of pre-STN and post-STN are $4$ and $1$, respectively.

\begin{figure}
    \centering
    \includegraphics[width=12cm]{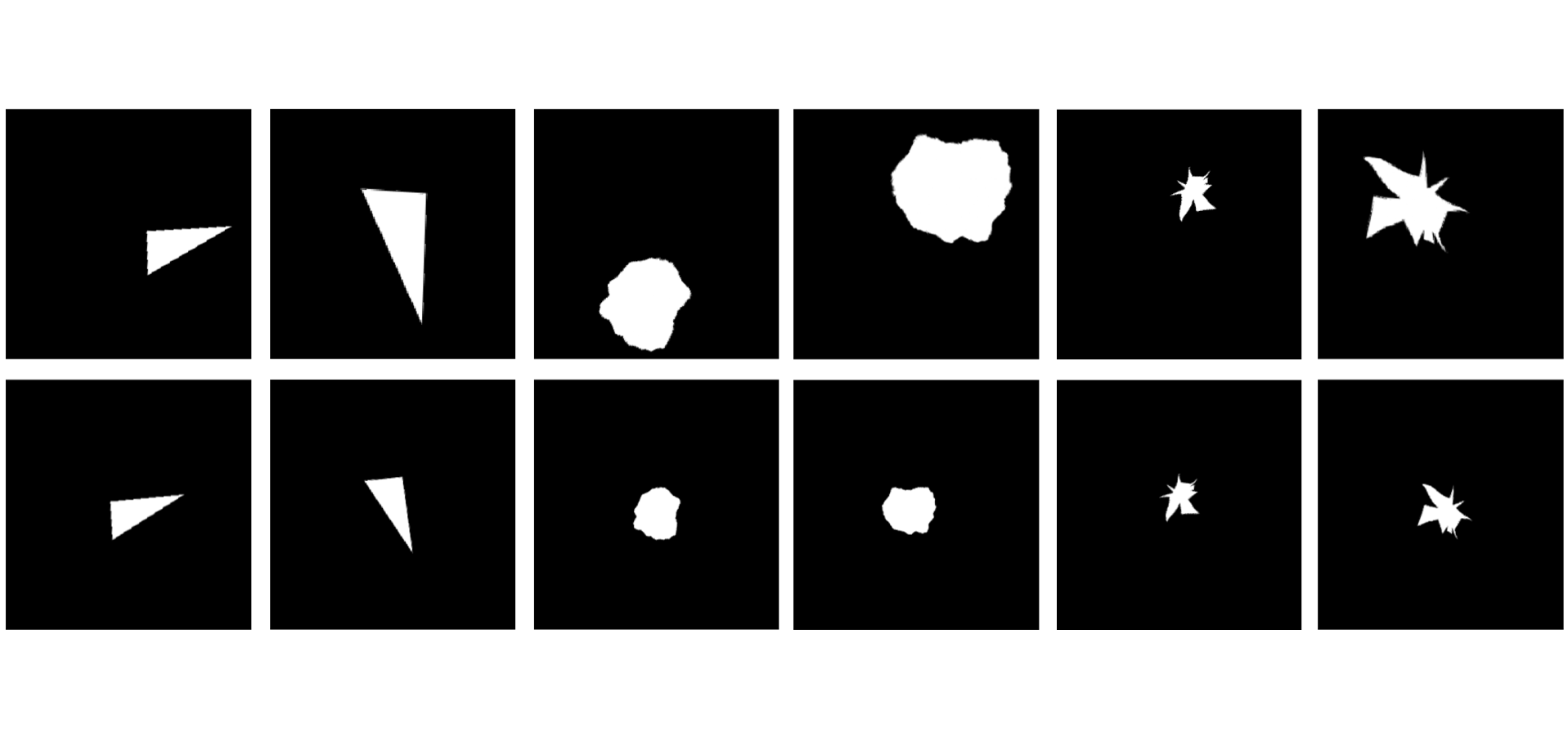}
    \caption{Illustration of the pre-STN effect. The first row shows input images with varying shapes, scales, positions, and orientations; the second row shows the corresponding pre-STN outputs after normalization.}
    \label{fig: prestn}
\end{figure}

The pre-STN is designed to assist the backbone in achieving the invariance property of HBS under translation, scaling, and rotation. It adjusts the size and orientation of shape $\Omega$ inside $I$ and puts it in the center of $I'$ by
\begin{equation}\label{eq: pre STN}
    \begin{pmatrix}
        x_i^s \\
        y_i^s
    \end{pmatrix} = \begin{pmatrix}
        k \cos\theta  & k \sin\theta & dx \\
        -k \sin\theta & k \cos\theta & dy
    \end{pmatrix} \begin{pmatrix}
        x_i^t \\
        y_i^t \\
        1
    \end{pmatrix},
\end{equation}
where $\phi^{preloc} = (dx, dy, k, \theta) \in \R^4$ is the location parameter of pre-STN. \cref{fig: prestn} clearly shows how the pre-STN works.

\begin{figure}
    \centering
    \includegraphics[width=9cm]{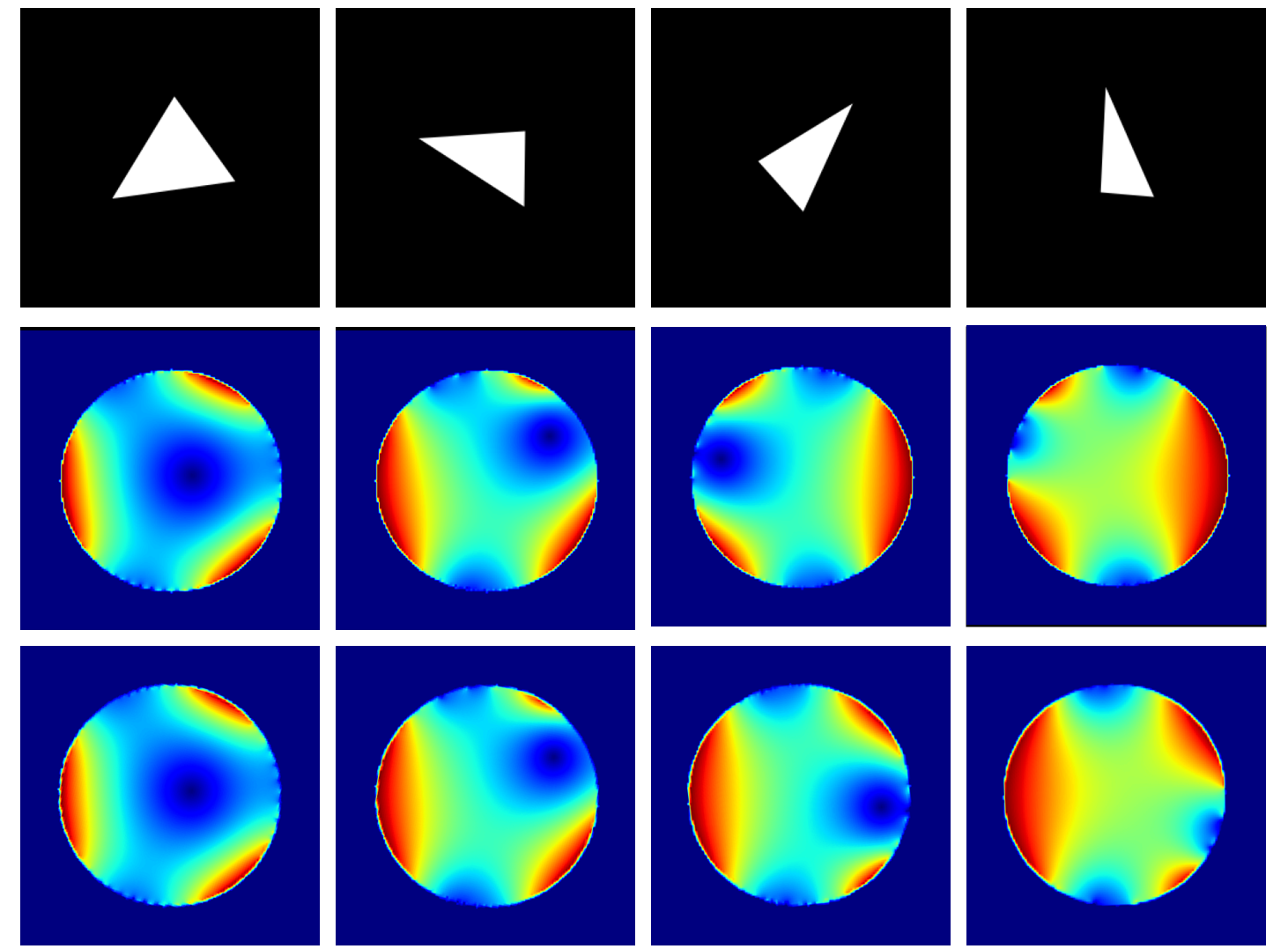}
    \caption{The illustration of angle inconsistency of HBS and post-STN. The 1st row comprises different images, the 2nd row is their HBS, and the 3rd row is the corresponding HBS normalized by the post-STN.}
    \label{fig: poststn}
\end{figure}

Meanwhile, the post-STN only normalizes the rotation of the output HBS $B'$ of the backbone. Angle inconsistency is a pain point for HBS in the application. The HBS of similar shapes, as the triangles shown in \cref{fig: poststn}, is generally similar, but as the difference increases, the HBS may have an obvious rotation, which may confuse the network in training and underestimate the shape similarity in inference. The post-STN effect $B'$ by
\begin{equation}\label{eq: post STN}
    \begin{pmatrix}
        x_i^s \\
        y_i^s
    \end{pmatrix} = \begin{pmatrix}
        \cos\theta & \sin\theta \\
        \sin\theta & \cos\theta
    \end{pmatrix} \begin{pmatrix}
        x_i^t \\
        y_i^t \\
    \end{pmatrix},
\end{equation}
where $\phi^{postloc} = \theta \in \R$ is the location parameter of post-STN.

\subsection{Loss function}
An appropriate loss function is essential for training neural networks. Our HBSN is a supervised network, so the kernel term of the loss function is the distance between the predicted HBS $B = N_\theta(I)$ and the reference HBS $\bar{B}$. To reduce the side effects of the angle inconsistent of HBS, $\bar{B}$ also needs to be fed into the post-STN, and the angle normalized reference $N_\theta^{post}(\bar{B})$ is adopted in following computing. The distance between two HBS $B$ and $B'$ can be determined by $L^2$ norm as
\begin{equation}\label{eq: d hbs}
    d(B, B') =  \frac{1}{N} {\sum_{(x, y) \in \D} \left(B(x, y)-B'(x, y) \right)^2},
\end{equation}
where $N = 7845$ is the number of pixels inside the unit disk under our setting.
Note that we only consider the differences inside the unit disk to reduce computational complexity and avoid unnecessary gradient backpropagation.
Therefore, the HBS loss $L_{HBS}$ is defined by
\begin{equation}\label{eq: hbs loss}
    L_{HBS}(I, \bar{B}, \theta) = d(N_\theta({I}) ,N_\theta^{post}(\bar{B})).
\end{equation}

\begin{figure}
    \centering
    \includegraphics[width=10cm]{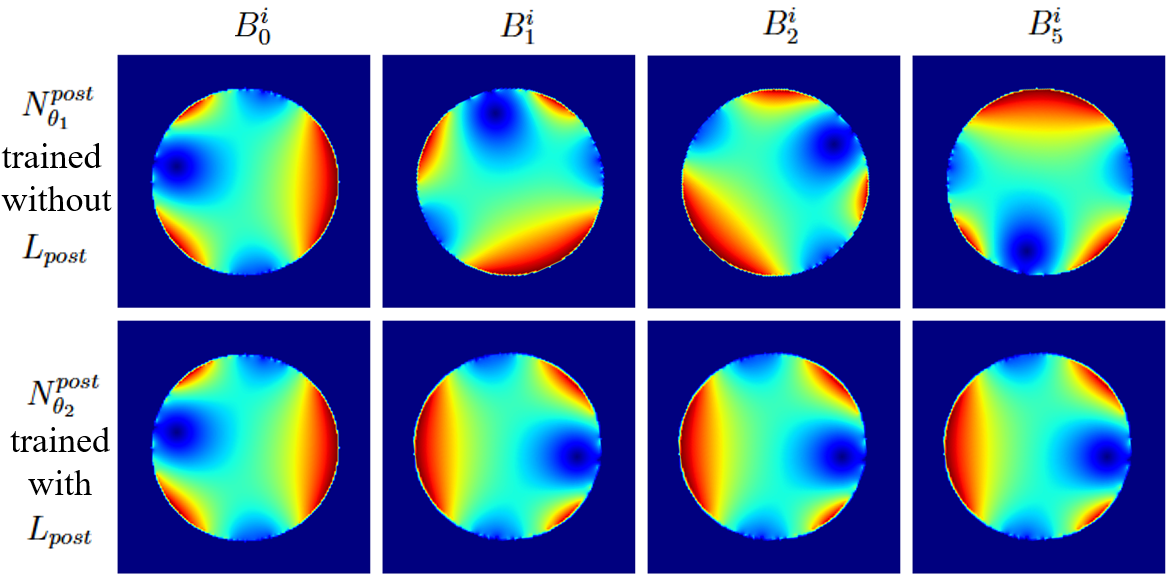}
    \caption{The comparison of effects of the post-STN of two networks trained with or without post-STN loss $L_{post}$. The 1st row shows unstable post-STN $N_{\theta_1}^{post}$ trained without $L_{post}$, while the 2nd row show $N_{\theta_2}^{post}$ trained with $L_{post}$. For $i$-row ($i=1, 2$), the 1st column is original HBS $B^i_0$, and the other 3 columns are $B^i_1$, $B^i_2$, and $B^i_5$, where $B^i_k = N_{\theta_i}^{post}(B^i_{k-1})$ is achieved by running post-STN $N_{\theta_i}^{post}$ for $k$ times.}
    \label{fig: with or without stn loss}
\end{figure}

However, such post-STN may fail to give the desired normalized HBS. We hope any output $B = N^{post}_\theta(B')$ is a fixed point under $N^{post}_\theta$, which means $B = N^{post}_\theta(B) $.
But if only using $L_{HBS}$, the post-STN may rotate any given HBS and never yield a stable result no matter how many times we run it, as shown in \cref{fig: with or without stn loss}. So we define the post-STN loss $L_{post}$ as
\begin{equation}\label{eq: post loss}
    L_{post}(I, \theta) = d(N_\theta^{post} \circ N_\theta(I), N_\theta(I)),
\end{equation}
to constrain the behavior of post-STN.

Finally, the whole loss function  of the proposed network is
\begin{equation}\label{eq: HBSN loss}
    L(I, \bar{B}, \theta) = L_{HBS}(I, \bar{B}, \theta) + \lambda_{post}L_{post}(I, \theta),
\end{equation}
where $\lambda_{post} \ge 0$ is the weight of post-STN loss.

\subsection{Integration with other segmentation models}\label{sec: combine hbsn}
\begin{figure}
    \centering
    \includegraphics[width=\textwidth]{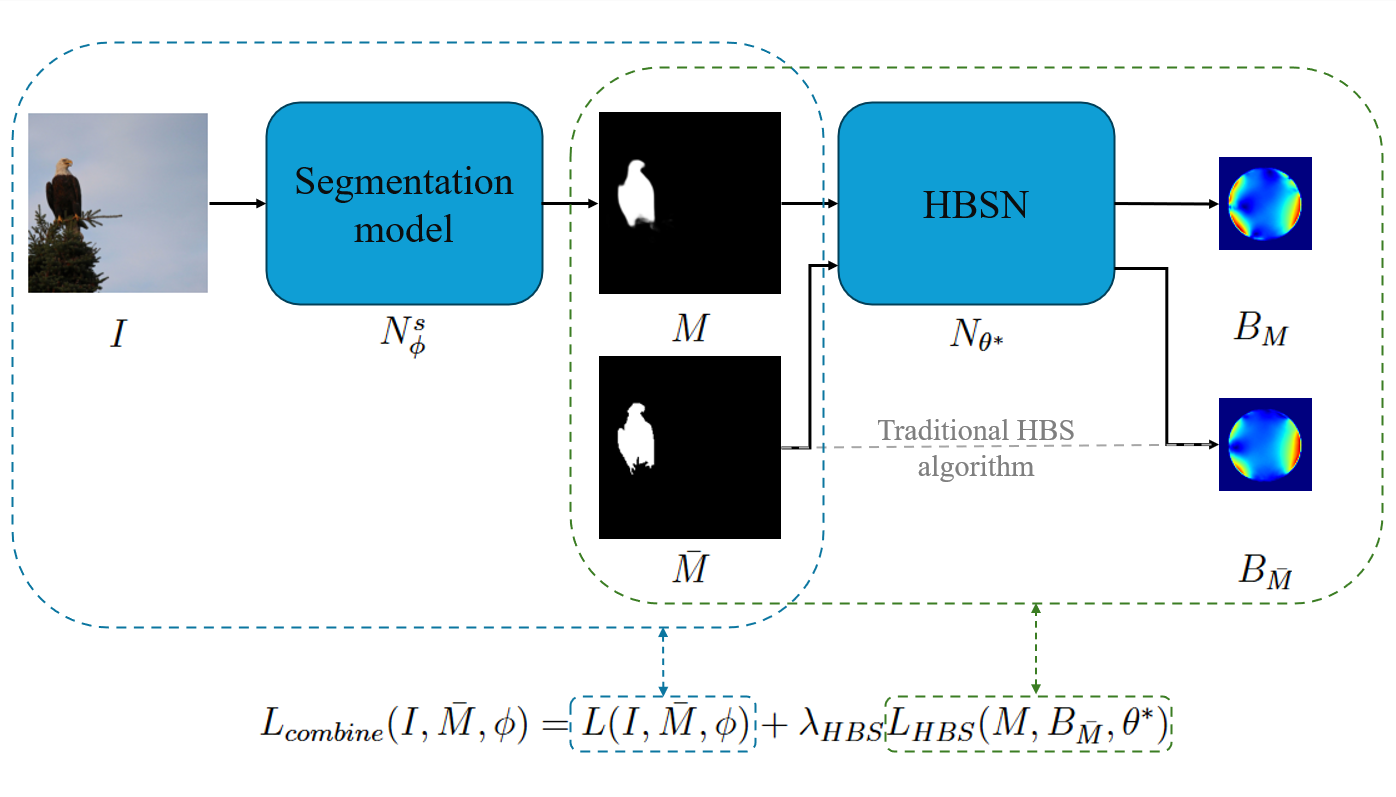}
    \caption{The illustration of integrating HBSN into other deep learning segmentation models.}
    \label{fig: combine hbsn}
\end{figure}
The primary purpose of HBSN is to provide a general method for introducing shape priors into deep learning segmentation networks, without modifying their architectures. Given a pretrained HBSN $N_{\theta^*}$, we describe how it can be used to supply shape prior information to other segmentation models. This method applies when the segmentation model is supervised and each image contains a single simply connected target object.

For any segmentation model $N^s_\phi$ with network parameters $\phi$, a mask $M$ is required as the network output to represent the segmentation result regardless of its detailed network structure. In general, values of such output mask $M$ range in $[0, 1]$, where the high-value area means the target object while the rest is the background. With proper training to $N^s_\phi$, the $M$ is almost binary and an ideal HBSN input.

The predicted mask $M$ of $N_\phi^s$ as well as the ground truth mask $\bar{M}$ could be fed into HBSN $N_{\theta^*}$ together then get their HBS $B_M$ and $B_{\bar{M}}$. Also, we could calculate all the $B_{\bar{M}}$ in advance to reduce the amount of computation when training.
The $L_{HBS}$ as \cref{eq: hbs loss} indicates the differences between $M$ and $\bar{M}$ given HBS.
Suppose the original loss function of model $N^s_\phi$ is $L(I, \bar{M}, \phi)$, new network parameters $\phi^*$ could be achieved after following optimization converges,
\begin{equation}\label{eq: plugable loss}
    \min_\phi L_{combine}(I, \bar{M}, \phi) = L(I, \bar{M}, \phi) + \lambda_{HBS} L_{HBS}(M, B_{\bar{M}}, \theta^*),
\end{equation}
where $\lambda_{HBS} \ge 0$ is the weight of HBSN loss. Therefore, we have achieved the integration of HBSN into any network for training or finetuning.

\section{Experimental result}\label{section: exp}
\begin{figure}
    \centering
    \includegraphics[width=\textwidth]{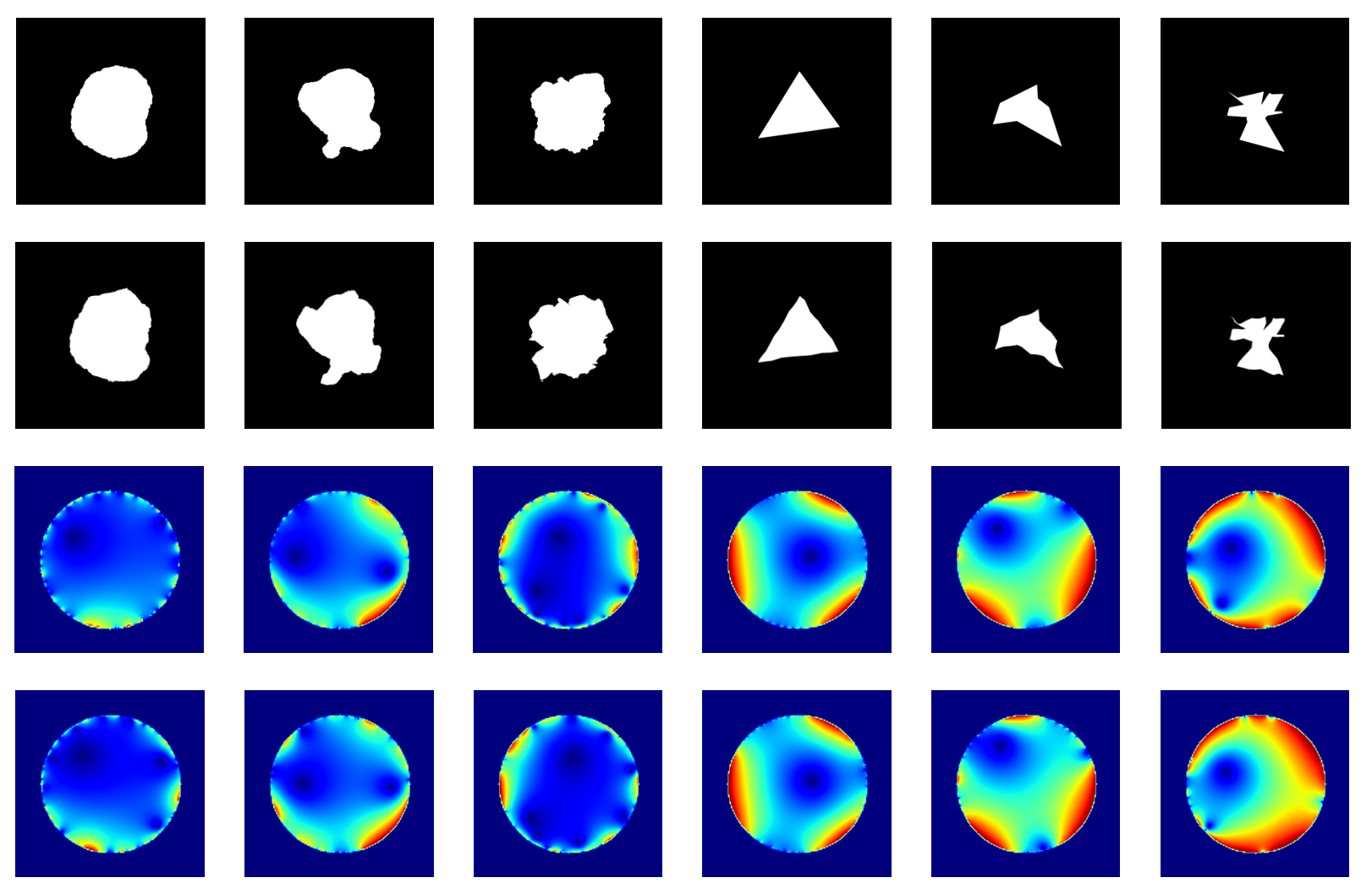}
    \caption{Examples of training data. Row 1: images from conformal welding (columns 1--3) and random polygon (columns 4--6) methods. Row 2: grid-perturbed versions. Rows 3--4: corresponding HBS.}
    \label{fig: hbsn training data}
\end{figure}
\subsection{Training data and augmentation}
We prepared about 20k simply connected shapes with their HBS for training. Shapes are generated via two methods: the \textbf{conformal welding method}, which reconstructs shapes from strictly monotone increasing functions $f: [0, 2\pi] \to [0, 2\pi]$ via the reverse Zipper algorithm\cite{marshall2012conformal,marshall2007convergence}, and the \textbf{random polygon method}, which connects $n$ random points clockwise to form simply connected regions. Grid perturbations are applied to enlarge the dataset (see \cref{fig: hbsn training data}).

For augmentation, translation, scaling, and rotation are applied randomly since they preserve HBS. To handle binary-like inputs during inference, we soften training images: given a binary image $I$ as \cref{eq: image with shape}, the softened image $I_s$ is
\begin{equation}\label{eq: soften}
    I_s(x, y) = \begin{cases}
         & \max \{ 1 - a + N(x, y)), 0.5\}, \; (x, y) \in \Omega,                  \\
         & \min \{b + N(x, y), 0.5-\epsilon \}, \; (x, y) \in D \setminus \Omega ,
    \end{cases}
\end{equation}
where $a, b \in [0, 0.2]$ are random offsets, $\epsilon > 0$ is a small positive number, and $N$ is Gaussian noise. Since $\Omega = \{(x, y) \mid I_s(x, y) \ge 0.5\}$ remains unchanged, the softened image preserves the same HBS.

\subsection{Evaluation metrics}
In this section, we report quantitative indicators and representative outputs of the trained HBSN to demonstrate its reliability in computing HBS from binary-like images. As defined in \cref{eq: d hbs}, the discrepancy between two HBS fields $B$ and $B'$ is measured by $d(B,B')$. Accordingly, we evaluate our model using the training objective in \cref{eq: HBSN loss},
\(
L = L_{HBS} + \lambda_{post} L_{post},
\)
where $L_{HBS}$ measures the prediction error in HBS space and $L_{post}$ regularizes the post-STN to produce a stable, rotation-normalized output.

\begin{figure}
    \centering
    \includegraphics[width=12cm]{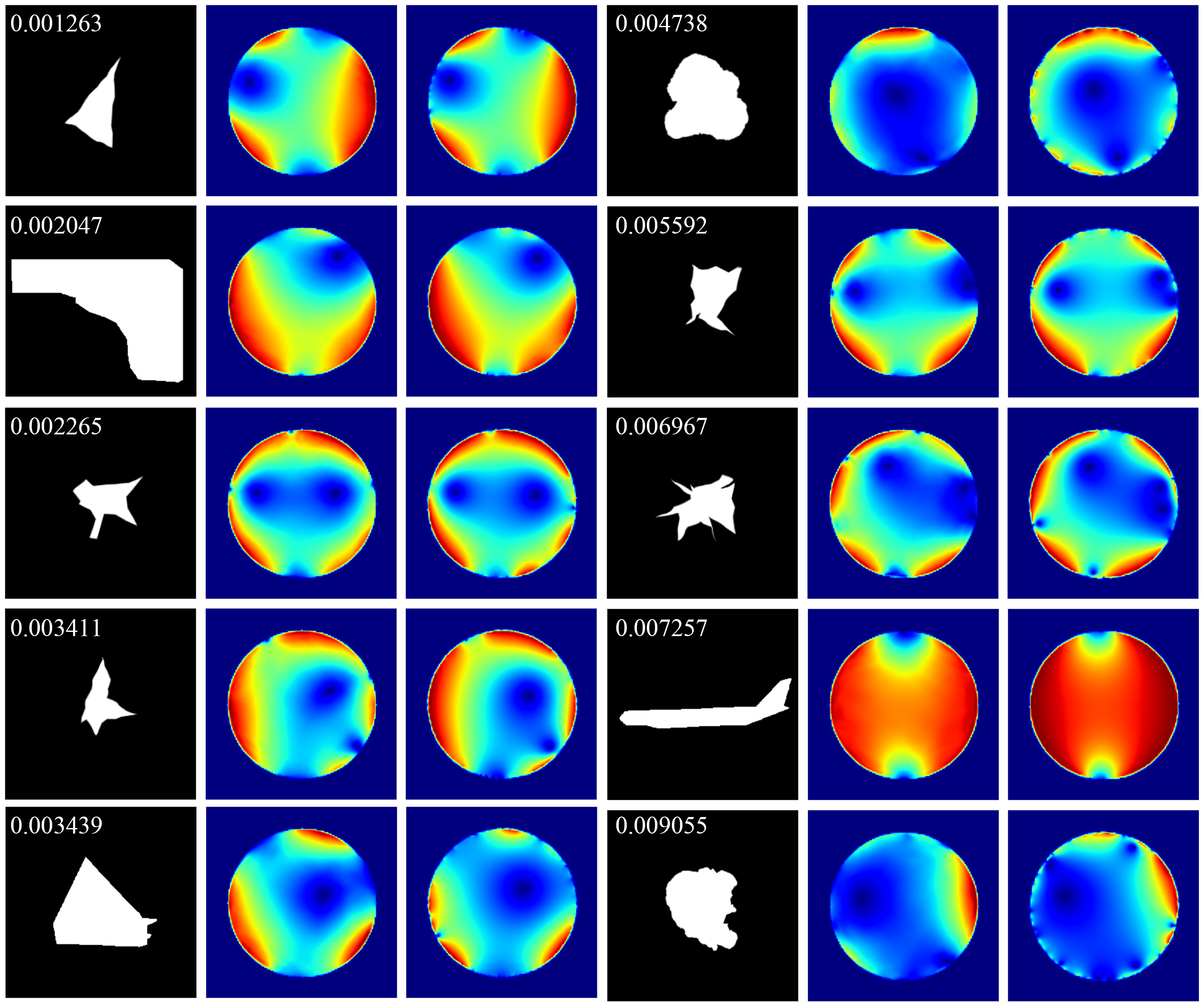}
    \caption{10 groups of randomly selected results of the best HBSN. Each group of results contains three images: the left is the input image, the middle is the predicted HBS, and the right is the ground truth. They are sorted in ascending order of $L_{HBS}$, which is marked in the left top of input images for each group, from top to bottom and left to right.}
    \label{fig: hbsn results}
\end{figure}

With $\lambda_{post}=0.1$, the best trained model achieves an average validation loss of $\overline{L_{HBS}}=0.006237$, indicating that most predictions are very close to the corresponding ground truth.
We also present a random selection of results from the validation set in \cref{fig: hbsn results}. Images 2, 5, and 9 are from the COCO dataset; the rest are synthetic. The predicted HBS are visually close to the ground truth in all cases, including those with above-average error.

\subsection{Architecture of backbone}
The best model (see \cref{fig: hbsn backbone}) downsamples five times with channels $[8,8,16,32,64,128]$ and upsamples four times with channels $[128,64,32,16,8]$. We evaluate various architectures based on accuracy ($L_{HBS}$), speed, and parameters in Table \ref{tab: backbone}. The best model achieves the highest accuracy while being hundreds of times faster than the traditional algorithm.

We investigate the influence of depth, channel size, and symmetry. \textbf{Depth:} Shallower networks (row 2) run faster but sacrifice accuracy, while deeper ones (row 3) increase training difficulty without improving performance. \textbf{Channel size:} Varying channel configurations (rows 4--6) failed to outperform the baseline. \textbf{Symmetry:} A symmetric decoder (row 7) with $256 \times 256$ output offers no advantage due to increased empty space outside the unit disk.

\begin{table}[]
    \centering
    \begin{tabular}{|c|c|c|c|}
        \hline
        Architecture & $L_{HBS}$         & \makecell{processing time            \\ for one image} & Parameters \\
        \hline
        \makecell{$[8,8,16,32,64,128]$                                          \\
            $\& [128,64,32,16,8]$}
                     & \textbf{0.006248} & $2.03 \pm 0.156$ ms
                     & 454,962                                                  \\
        \hline
        \makecell{$[8,8,16,32,64]$                                              \\
            $\& [64,32,16,8]$}
                     & 0.01197           & $\boldsymbol{1.81 \pm 0.230}$ ms
                     & \textbf{114,162}                                         \\
        \hline
        \makecell{$[8,8,16,32,64,128,256]$                                      \\
            $\& [256,128,64,32,16,8]$}
                     & 0.007515          & $3.02 \pm 0.159$ ms
                     & 1,816,498                                                \\
        \hline
        \makecell{$[8,16,32,64,128,256]$                                        \\
            $\& [256,128,64,32,16]$}
                     & 0.006838          & $2.44 \pm 0.118$ ms
                     & 1,813,434                                                \\
        \hline
        \makecell{$[8,8,8,16,32,64]$                                            \\
            $\& [64,32,16,8,8]$}
                     & 0.008522          & $1.96 \pm 0.115$ ms
                     & 116,854                                                  \\
        \hline
        \makecell{$[8,8,16,32,64,256]$                                          \\
            $\& [256,64,32,16,8]$}
                     & 0.007247          & $2.28 \pm 0.158$ ms
                     & 1,033,074                                                \\
        \hline
        \makecell{$[8,8,16,32,64,128]$                                          \\
            $\& [128,64,32,16,8,8]$}
                     & 0.006954          & $2.17 \pm 0.180$ ms
                     & 456,470                                                  \\
        \hline
        traditional algorithm
                     & 0                 & $871 \pm 23.5$ ms                & - \\
        \hline
    \end{tabular}
    \caption{Performance of network in different architectures.}
    \label{tab: backbone}
\end{table}

\begin{table}[]
    \centering
    \begin{tabular}{|c|c|c|}
        \hline
        $L_{HBS}$    & w/ pre-STN        & w/o pre-STN \\
        \hline
        w/ post-STN  & \textbf{0.006237} & 0.006557    \\
        \hline
        w/o post-STN & 0.007597          & 0.008721    \\
        \hline
    \end{tabular}
    \caption{Performance with or without pre-STN and with or without post-STN.}
    \label{tab: stn}
\end{table}
\subsection{Performance of pre- and post-STN}
As shown in \cref{fig: prestn}, the pre-STN adjusts the size, position, and orientation of input images. For position, it centers all shapes within the image, as expected. For size, all shapes are scaled to a smaller size; this follows from the input resolution of $256 \times 256$, the output resolution of $128 \times 128$, and the unit disk mask radius of $50$ pixels. For orientation, the pre-STN applies only small rotations, consistent with the CNN's approximate rotational invariance and the post-STN's role in handling residual angle differences.

The post-STN behavior is illustrated in \cref{fig: poststn} and \cref{fig: with or without stn loss}. It consistently maps similar HBS representations to nearly the same angle, and this angle remains stable under repeated application.

In addition, to ensure the stability of the post-STN, an additional term $L_{post}$ with a weight $\lambda_{post} = 0.1$ is added to the loss function. Actually, $L_{post}$ is about two orders of magnitude smaller than $L_{HBS}$, indicating that the training process is still primarily dominated by the differences between the predicted HBS and the reference HBS. However, the post-STN still achieves the desired effect, so further discussion on the value of $\lambda_{post}$ is not pursued.

The pre-STN and post-STN modules are not tightly coupled with the backbone, allowing for flexibility in their usage. The following experiments trained networks using the best backbone architecture under four conditions: with or without pre-STN and with or without post-STN. Performance is evaluated on the validation set under each condition and displayed in \cref{tab: stn}. The table confirms that both modules contribute to improved accuracy, with the post-STN having a larger effect.

\begin{figure}
    \centering
    \includegraphics[width=13cm]{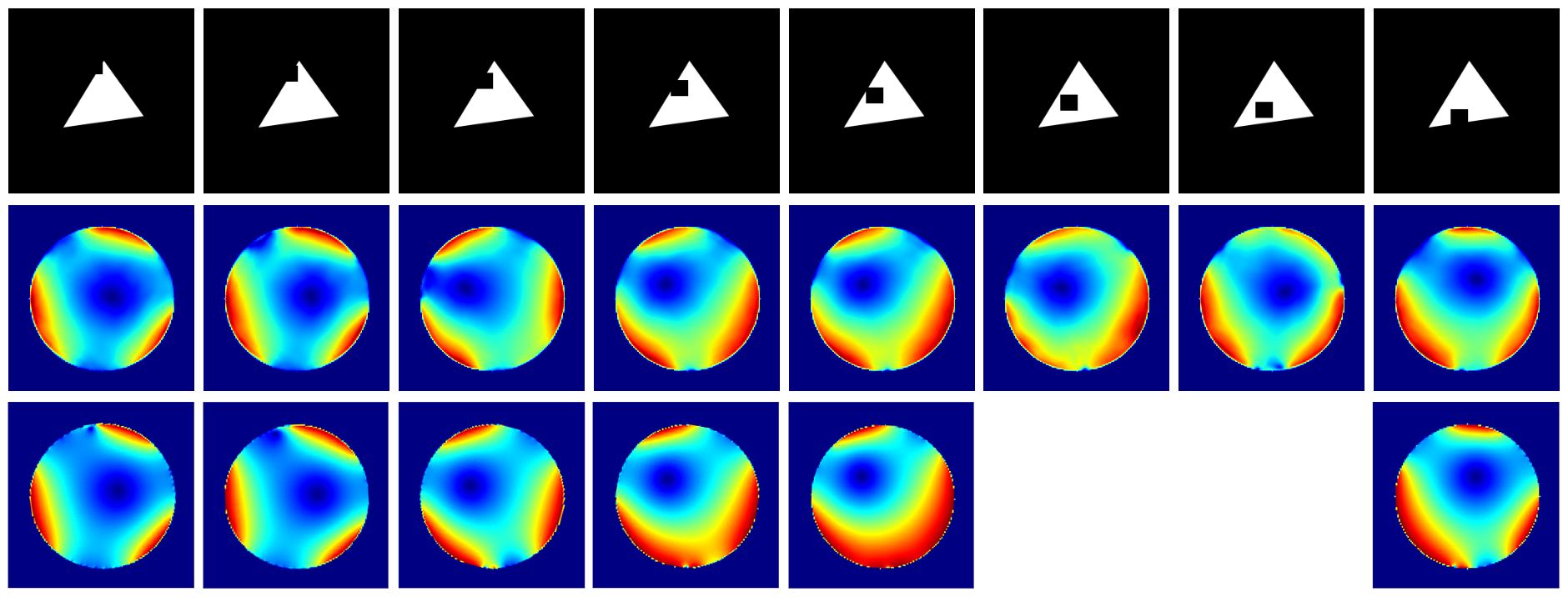}
    \caption{Results of HBSN on multi-connected shapes. The first row is given images, the second row is HBSN predictions, and the last row is HBS computed by traditional algorithm, where the 6th and 7th images are multi-connected and don't have HBS.}
    \label{fig: hbsn on mutilconnected}
\end{figure}

\begin{figure}
    \centering
    \includegraphics[width=13cm]{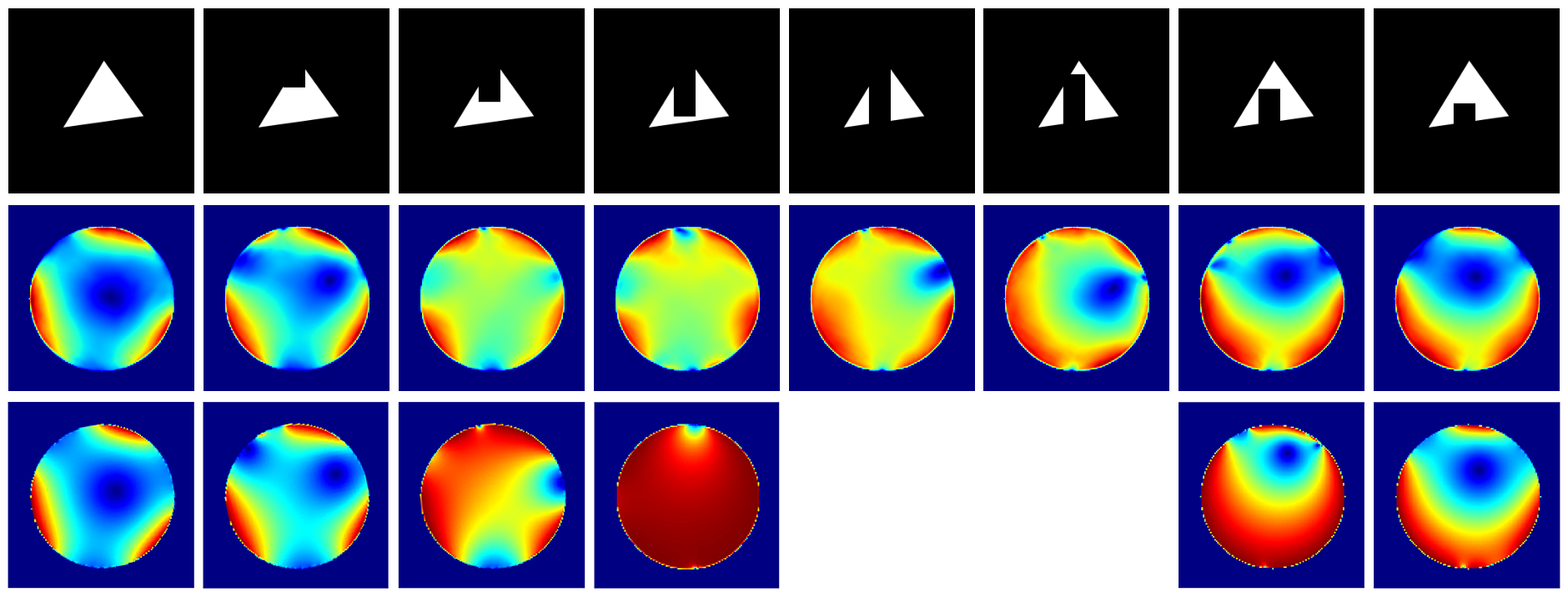}
    \caption{Results of HBSN on disconnected shapes. The first row is given images, the second row is HBSN predictions, and the last row is HBS computed by traditional algorithm, where the 5th and 6th images are disconnected and don't have HBS.}
    \label{fig: hbsn on disconnected}
\end{figure}

\subsection{HBSN on disconnected or multi-connected shapes}
HBS is defined only for simply connected shapes, while multi-connected or disconnected regions have no valid HBS. But during training of segmentation networks under framework as \cref{fig: combine hbsn}, intermediate predictions may temporarily violate this assumption. Thus, we test a pretrained HBSN on such inputs in \cref{fig: hbsn on mutilconnected,fig: hbsn on disconnected}.

We occlude a triangle with a small square to generate a series of images: the shape is multi-connected when the square is fully contained within the triangle, and simply connected otherwise. As shown in \cref{fig: hbsn on mutilconnected}, HBSN predicts accurately for simply connected cases and yields reasonable results via an interpolation-like method for multi-connected ones (6th–7th columns). Using a long narrow rectangle to occlude a triangle produces disconnected ones shown in \cref{fig: hbsn on disconnected}, where HBSN significantly underestimates shape distortion even for simply connected cases (e.g., columns 3, 4, 7), likely due to the absence of such images in the training set, yet it still predicts similar HBS for similar images. HBSN's output remains stable for disconnected and multi-connected inputs without degrading training, suggesting a possible direction for extending the HBS framework to a broader class of shapes.

\subsection{Plug-and-play enhancement of segmentation networks}
\begin{table}[]
    \centering
    \begin{tabular}{|c|c|c|}
        \hline
        Models         & Dice   & IoU    \\
        \hline
        UNet           & 0.7747 & 0.7008 \\
        \hline
        UNet+HBSN      & 0.7858 & 0.7143 \\
        \hline
        DeepLabV3      & 0.7630 & 0.6826 \\
        \hline
        DeepLabV3+HBSN & 0.7757 & 0.6958 \\
        \hline
    \end{tabular}
    \caption{Performance on COCO val2017 subset.}
    \label{tab: combine hbsn}
\end{table}

We evaluate whether a pretrained HBSN can serve as a generic plug-in module following \cref{sec: combine hbsn}. We train UNet \cite{ronneberger2015u} and DeepLabV3 \cite{chen2017rethinking} on COCO train2017 and evaluate on COCO val2017, filtering images to contain a single simply connected object (area $>100$ pixels) so that the ground-truth mask admits a well-defined HBS.

\begin{figure}
    \centering
    \includegraphics[height=20cm]{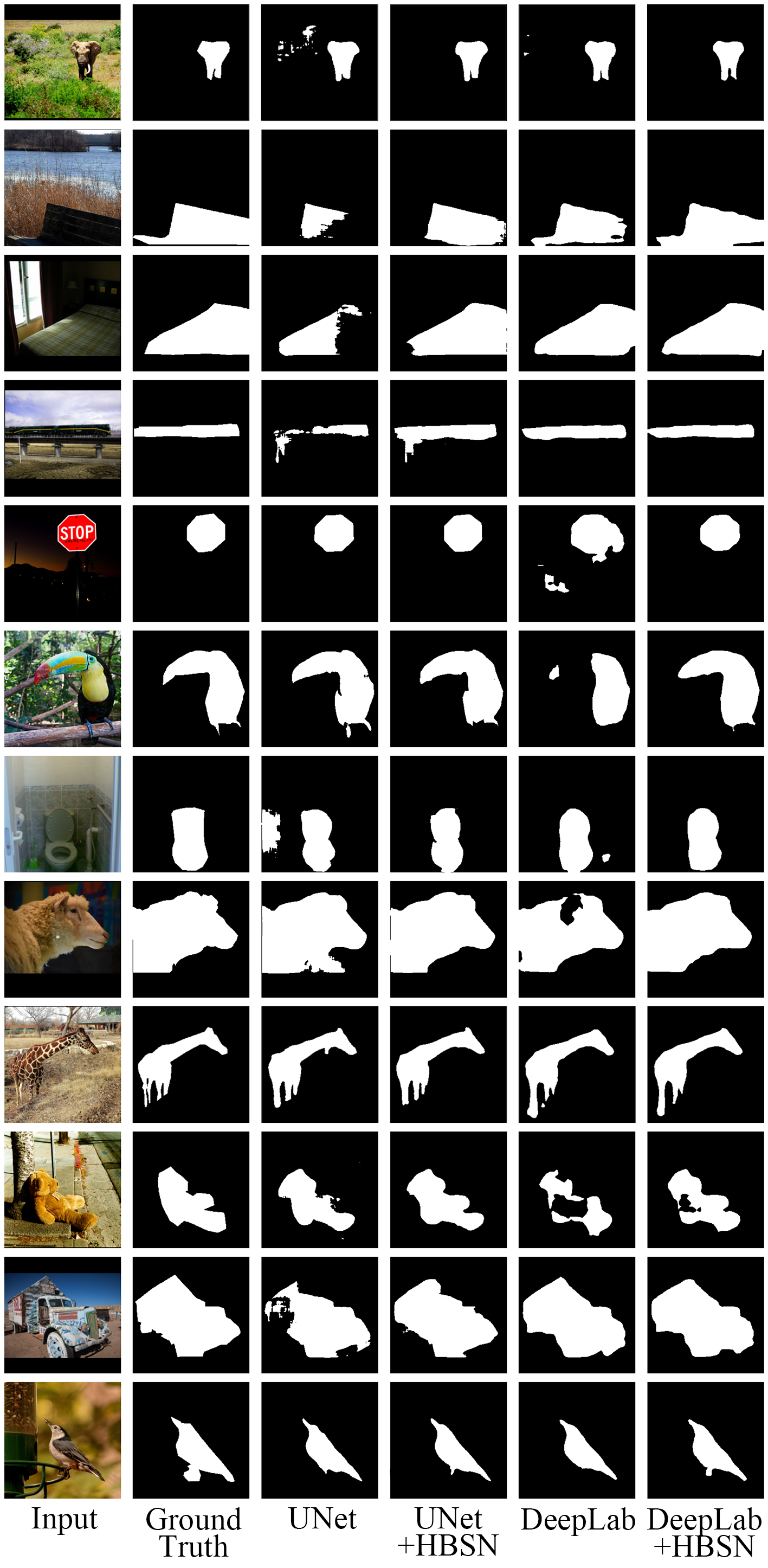}
    \caption{Segmentation results on COCO val2017 subset.}
    \label{fig: coco seg results}
\end{figure}

As shown in \cref{tab: combine hbsn,fig: coco seg results}, adding the HBSN-based loss consistently improves IoU and Dice for both baselines without modifying their architectures. The pretrained HBSN parameters remain fixed, making integration straightforward. Geometrically, pixel-space losses encourage local consistency but may under-penalize structured boundary errors at reasonable overlap. The HBS-based term instead compares shape information, injecting a long-range constraint that promotes more complete masks with fewer missing parts. Since HBS is invariant under translation, scaling, and rotation, the HBS loss focuses on intrinsic shape discrepancies rather than position or scale differences. HBSN is designed for supervised single-object settings and should be understood as a refinement signal that complements, rather than replaces, the visual feature learning of the base network.

\subsection{HBS-space discrepancy as a complementary supervision signal}

\begin{figure}
    \centering
    \includegraphics[height=16cm]{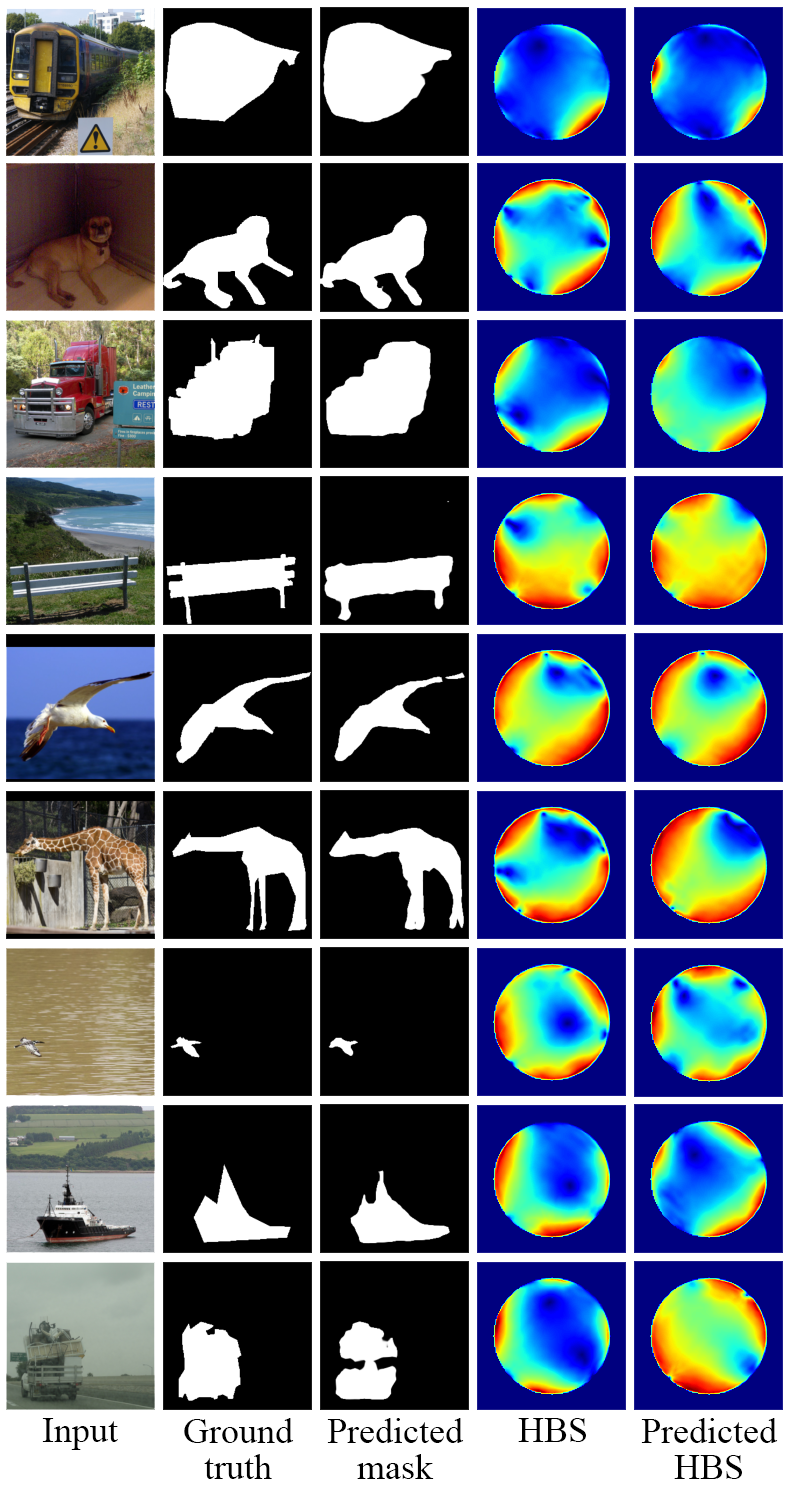}
    \caption{Visualization of HBS-space discrepancy during UNet+HBSN training. Although the predicted and ground truth masks appear visually similar, their HBS representations exhibit noticeable differences, demonstrating that HBS captures geometric boundary errors not fully reflected by pixel-space overlap metrics.}
    \label{fig: helps of shape prior}
\end{figure}

We further investigate how the HBSN provides complementary shape information during segmentation training to improve the performance of original networks. We select several snapshots from the training process of UNet+HBSN and visualize them in \cref{fig: helps of shape prior}. At the displayed stage, the model attains an IoU of approximately $0.82$ on the training set. From left to right, each row reports the input image $I$, the ground truth mask $\bar{M}$, the predicted mask $M$, the ground truth HBS $B_{\bar{M}}$, and the predicted-mask HBS $B_M$, where all HBS are computed by the same pretrained HBSN. Although $M$ and $\bar{M}$ appear visually close and differ mainly in fine boundary details, their HBS can still differ substantially.

This phenomenon suggests that the discrepancy in HBS space captures geometric errors that are not fully reflected by pixel-space metrics. Intuitively, IoU/Dice are dominated by area overlap and can become insensitive once predictions reach a reasonably good overlap, especially when the remaining errors concentrate on thin boundary regions. In contrast, HBS summarizes global conformal distortion induced by the shape boundary and is invariant to translation, scaling, and rotation; hence it focuses the learning signal on intrinsic geometry rather than nuisance pose variations. Small but spatially coherent boundary deviations (e.g., missing protrusions, over-smoothing of corners, or local leakage that changes curvature patterns) may lead to a noticeable change of the conformal welding and thus a non-negligible change in the HBS. Therefore, incorporating an HBS-based loss effectively ``amplifies'' these residual geometric discrepancies and provides a long-range regularization that couples distant boundary parts through a global descriptor. In practice, this complementary signal is particularly helpful when the segmentation network already produces masks with good overlap but still lacks geometric fidelity, and it explains why HBSN-guided training can further refine the predicted shapes toward the ground-truth geometry.

\section{Conclusion}\label{section: conclusion}
This paper presents the Harmonic Beltrami Signature Network (HBSN), a deep learning architecture for computing the Harmonic Beltrami Signature (HBS) from binary-like images. By exploiting the function approximation and differentiability of neural networks, HBSN enables efficient extraction of shape prior information. Experiments confirm that HBSN accurately predicts HBS representations.

The computational efficiency and differentiable structure of HBSN make it well-suited for integration with existing segmentation frameworks. This allows end-to-end training of segmentation models that incorporate HBS-based shape priors, improving their accuracy and robustness.

Several directions remain open for future work. Extending HBSN to handle multiple shape priors or combinations of shape representations could improve its generality and robustness. Applying HBSN to real-time segmentation tasks---such as video object segmentation or augmented reality---would extend its use to dynamic settings. Evaluating HBSN across diverse datasets and domains would also clarify its potential for transfer learning and domain adaptation.

\bibliographystyle{plain}
\bibliography{paper}

\end{document}